\documentclass[10pt,twocolumn,letterpaper,hyperref={pagebackref,breaklinks,colorlinks,allcolors=cvprblue}]{article}

\usepackage[pagenumbers]{cvpr} 
\usepackage{orcidlink} 

\usepackage{cuted} 

\usepackage{currfile} 

\usepackage{subcaption}
\usepackage{graphicx}
\usepackage{color}
\usepackage{tabularray}
\UseTblrLibrary{booktabs}
\definecolor{Silver}{rgb}{0.8,0.8,0.8}
\usepackage{makecell}
\usepackage{rotating}
\usepackage{pgfplots}
\pgfplotsset{compat=1.18}
\usepackage{longfigure}

\definecolor{cvprblue}{rgb}{0.21,0.49,0.74}

\title{Counter-DECEIVR: Counterfactual Explanations for Deepfake Detectors using Detector-Guided Rectified Flow Inversion\thanks{This work was supported by EU Horizon Europe Program under Grant 101214398 ELLIOT.}}

\author{Mariano V. Ntrougkas\orcidlink{0009-0004-0569-0837}$^{1,2}$,
 Vasileios Mezaris\orcidlink{0000-0002-0121-4364}$^{1}$, and
 Ioannis Patras\orcidlink{0000-0003-3913-4738}$^{2}$\\
$^{1}$Information Technologies Institute / CERTH, Thessaloniki, Greece\\
$^{2}$Queen Mary University of London, London, UK}

\begin{document}
\maketitle
\begin{abstract}
Deep-learning-generated media are increasingly passed off as real, driving an arms race between image generators and deepfake detectors. Trust in the latter is limited by their often black-box nature; and, while region-based explanations can sometimes be extracted, these become less useful as generation quality improves. Counterfactual examples, which show the minimal change that flips a detector's decision, reveal what evidence drives a verdict, helping analysts decide whether to trust it. This work introduces Counter-DECEIVR (DEtector-guided reCtifiEd flow InVeRsion for counterfactuals), a novel training-free framework that generates counterfactual examples. Counter-DECEIVR steers latents in the velocity field of a rectified flow model using gradients from off-the-shelf deepfake detectors, without training a generator or adapting the detector; unlike prior methods, it requires no per-instance or per-detector fitting and no diffusion training on the target data. It also leverages the detector's explanation maps to concentrate editing to salient regions, turning low-level feature importance into localised counterfactual edits. A single parameter controls the strength of the edits. Across two deepfake detectors and two widely used forensics datasets, Counter-DECEIVR outperforms prior counterfactual generation methods, including training-based ones. Code will be made available upon acceptance. 
\end{abstract}    
\section{Introduction}
\label{sec:intro}
\begin{figure}[ht]
    \centering
    \includegraphics[width=0.9\linewidth]{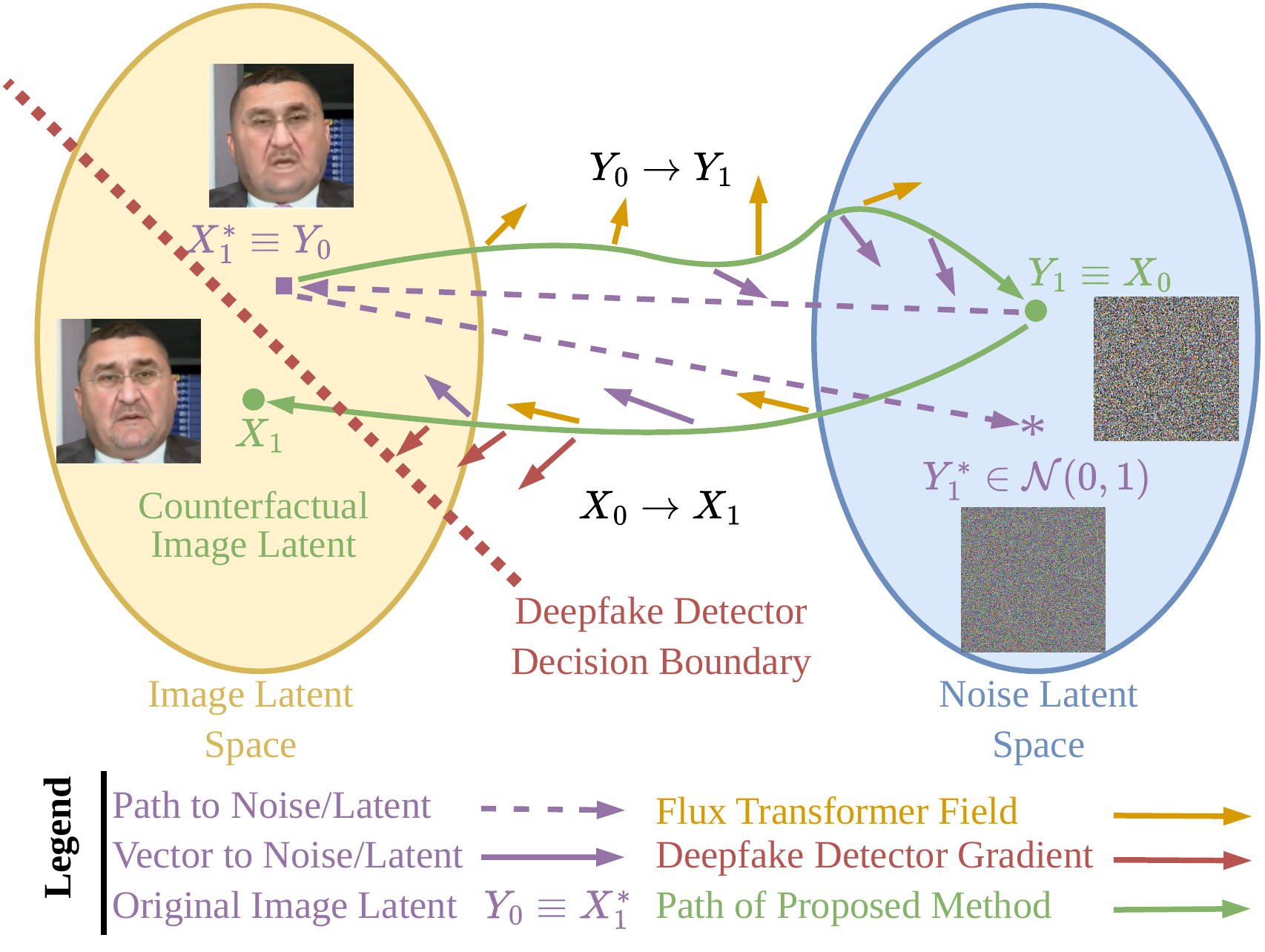}
    \caption{Counter-DECEIVR inverts a query image to noise with a rectified flow, then guides the reverse trajectory with the detector's own gradient, steering the image across the detector's decision boundary by affecting only the regions the detector relies on. $Y_1^* \sim \mathcal{N}(0,1)$ is a sampled Gaussian noise that acts as the target of the inversion path, and $X_1^* \equiv Y_0$ is the original image latent that acts as the target of the reverse path, keeping the counterfactual $X_1$ close to the query image.}
    \label{fig:general}%
\end{figure}

Generative AI methods can produce or alter media with increasingly high quality \cite{peideepfake2026}. There is significant societal harm associated with presenting such generated content as originating from the real world and using it to spread disinformation. This is especially the case for deepfakes: images and videos taken from the real world that have been altered using AI methods. This has created the need for tools, namely deepfake detectors, that end-users, such as news agencies, can employ to validate that received material is not manipulated or fabricated.

Deepfake detection models have entered into an arms race with generative AI methods \cite{peideepfake2026}, and manipulated images are often detected as such only from high-frequency artifacts that are indiscernible to humans \cite{tanfrequencyaware2024}. An explanation map that highlights these subtle artifacts is therefore of limited value \cite{pengcounterfactual2022}: it points to where the detector looked, but the evidence it points to is not something a human can inspect. This insufficiency motivates a search for different forms of explanation. Counterfactual explanations \cite{wachtercounterfactual2018} instead propose a minimal, salient, and plausible edit to the image: minimal, in that the edit is the smallest change that flips the prediction; salient, in that the change is perceptible and confined to the regions the model relied on; and plausible, in that the edited image remains a realistic image. Explanation maps show where the model found possibly imperceptible artifacts, while counterfactuals demonstrate what in those regions drives the decision, which is why they can be regarded as higher-order explanations \cite{pearlbook2018}.

Prior works generate counterfactual explanations by training per-instance representation networks (DISC \cite{thiagarajandesigning2021}), by classifier-guided diffusion (DiME \cite{jeanneretdiffusion2022}, ACE \cite{jeanneretadversarial2023}), or by fine-tuning Stable Diffusion on deepfake datasets \cite{yangunveiling2025}. Practically, all must be fitted to the target domain: DISC fits a network to every query image, DiME and ACE require a diffusion model trained on the target data, and the fine-tuned Stable Diffusion model must be retrained for each detector, a cost that recurs as detectors evolve. Methodologically, DDPM-based guidance struggles at high noise levels, where the one-step clean estimate is too blurry to yield a useful detector gradient: DiME compensates with costly nested denoising at every step, while ACE starts from shallow noise, confining edits to near-adversarial perturbations that flip the decision without revealing what changed.

Motivated by these limitations, we propose Counter-DECEIVR (DEtector-guided reCtifiEd flow InVeRsion for counterfactuals), a novel framework that combines rectified flow inversion \cite{routsemantic2024} and classifier-guided diffusion \cite{patelsteering2024} to generate counterfactuals from a single pretrained flow model and an off-the-shelf detector, with no per-instance or per-detector fitting and no target-data training. Two design choices drive its performance. First, rectified flow's near-straight trajectories make a one-step clean-image estimate sharp enough for detector guidance from high noise, by contrast to DiME's costly nested denoising and ACE's shallow, near-adversarial edits. Second, the detector's own explanation map weights the guidance, so edits concentrate on the regions most salient to the detector while the rest of the image is reconstructed. Together, these yield sparse, localised counterfactuals. On two detectors across FaceForensics++ \cite{rosslerfaceforensics2019} and Celeb-DF \cite{licelebdf2020}, Counter-DECEIVR flips decisions more efficiently at equal realism than state-of-the-art methods, which must alter more of the image and sacrifice more realism to do so (Fig.~\ref{fig:curves}), and its edits stay mostly confined to the regions the detector relies on (Fig.~\ref{fig:methods}).

In summary, our contributions are:
\begin{itemize}

  \item We propose Counter-DECEIVR, a training-free framework that steers a pretrained rectified flow model with the gradients of any differentiable deepfake detector to generate counterfactual examples, normalising the detector gradient to the magnitude of the flow step so that it acts as a superposed velocity.
  \item We show that the proposed rectified-flow-based scheme, owing to its near-straight trajectories, supports off-the-shelf detector guidance from high noise levels via a one-step clean-image estimate, without target-data training, by contrast to DDPM-based \cite{hodenoising2020} methods requiring nested denoising (DiME \cite{jeanneretdiffusion2022}) or shallow noise (ACE \cite{jeanneretadversarial2023}).
  \item Unlike prior methods, which ignore the detector's saliency, we use its explanation maps as a novel spatial prior that weights both the velocity field and the guidance step, yielding sparser, more localised counterfactuals and thus more informative explanations.
  \item We experimentally show that Counter-DECEIVR achieves a better trade-off between edit efficiency and realism than state-of-the-art methods on two deepfake detectors and two forensic benchmarks.
\end{itemize}
\section{Related Work}
\label{sec:related}
Current AI systems largely lack the capability to explain their decisions, which hinders user trust and adoption. Counterfactual examples, a form of explanation, are commonly classified according to Pearl's Ladder of Causation \cite{pearlbook2018}, a three-tier hierarchy of causal reasoning: association ``what if I see…?''), intervention (``what if I do…?''), and counterfactuals (``what if I had done…?''). Counterfactual reasoning occupies the highest level, asking what the outcome would have been under a hypothetical change to the input, holding everything else fixed. Formally, a counterfactual asks for $P(Y_{X=x'}=y\, |\,E=e)$, which should be read as: the probability that the outcome $Y$ would be $y$ if we set $X$ to $x'$, given the evidence $E=e$ that we actually observed. In other words, it compares the observed world with a hypothetical world in which one input variable is changed while the rest of the observed context is held fixed. For a classifier $f$, we must relax this definition, since we cannot intervene on the data-generating process and there is no ground-truth causal graph. We therefore treat the intervention as a constrained perturbation in image space, replacing the causal query with a search for a perturbation $x' = x + \delta$ that is minimal, salient, and plausible: it changes the predicted label, $f(x') \neq f(x)$, remains a plausible image, and is concentrated on the regions the detector used for its initial decision. This follows the definition given in \cite{wachtercounterfactual2018}, and as we show, how minimal a perturbation needs to be is itself a case-dependent choice. These three requirements distinguish counterfactual examples from explanation maps. Unlike explanation maps, they answer why a decision was made and what would change it, in a form a human can inspect. Without the salient, plausible perturbation requirement, this reduces to an adversarial attack at the testing phase \cite{chakrabortysurvey2021}, which seeks to force a targeted model to produce incorrect outputs rather than explain its decisions.

There are unique challenges to explaining the decisions of deepfake detectors that make explanation maps an ill-suited format. Artifacts may not be human-perceptible, which makes highlighting them less valuable. Additionally, there are non-localizable high-frequency traces of deepfakes that detectors increasingly rely on \cite{tanfrequencyaware2024}. Counterfactual examples thus provide a better-suited format for explaining these methods. Counterfactual generation for deepfake detectors remains underexplored; we have found only three notable prior methods that generate counterfactual examples for explaining deepfake detectors, namely the work of \cite{pengcounterfactual2022}, Artifact-Eraser \cite{licounterfactual2024} and the work detailed in \cite{yangunveiling2025}. The latter is trained with an ensemble, and its counterfactuals target no single detector, so they cannot be attributed to an individual model.

Prior counterfactual generation methods fulfill the plausibility requirement by modeling the target data manifold so the perturbed image remains on-distribution, in one of two ways. The first (DISC \cite{thiagarajandesigning2021}) trains an auxiliary error-prediction module \cite{yoolearning2019} on the target classifier's data and fits a per-instance  representation network such as INR \cite{sitzmannimplicit2020} or DIP \cite{ulyanovdeep2020} to reproduce each query image, and progressively unfreezes it towards the counterfactual objective. The jointly trained error-prediction module learns aspects of the target classifier's data distribution, supplying the plausibility requirement. The second trains a generative model on the target data. DiME \cite{jeanneretdiffusion2022} trains a diffusion model on the target data and guides it with the gradients of an existing classifier. ACE \cite{jeanneretadversarial2023}, like DiME, uses guided diffusion, but in two stages: first, it computes a counterfactual image via a classifier-guided noise diffusion process \cite{hodenoising2020}, then repeats the process using a mask of the regions changed most in the first stage \cite{lugmayrrepaint2022}. This second stage is the closest prior analogue to a saliency requirement, but it reflects where the diffusion process made the most significant edits, not where the detector looks. Artifact-Eraser \cite{licounterfactual2024} instead optimizes, for each query image, the latent code of a StyleGAN trained on the target data; we omit it from our comparisons, since it was already compared against the trained Stable Diffusion v1.4 model in \cite{yangunveiling2025}. The authors of \cite{yangunveiling2025} in turn train a Stable Diffusion v1.4 model on the FF++ \cite{rosslerfaceforensics2019}, DFDC \cite{dolhanskydeepfake2020}, and Celeb-DF(v2) \cite{licelebdf2020} using an ensemble of deepfake detectors, producing counterfactuals with no classifier signal at inference. The method of \cite{pengcounterfactual2022} falls outside of this dichotomy by directly optimizing a perturbation in pixel space and only indirectly encouraging plausibility by limiting the extent of the perturbation; It is thus closer to an adversarial attack than to a full counterfactual example, and we omit it as well. Counter-DECEIVR differs on all three requirements: The point at which classifier guidance is enabled controls how strong the edits will be, plausibility is inherited from a pretrained rectified flow model rather than fitted to the target data, and saliency is taken from the detector's own explanation map rather than from the diffusion trajectory.

\section{Methodology}
\label{sec:method}

\subsection{Overview}
\label{ssec:overview}
\begin{figure*}[ht]
    \centering
    \includegraphics[width=1\linewidth]{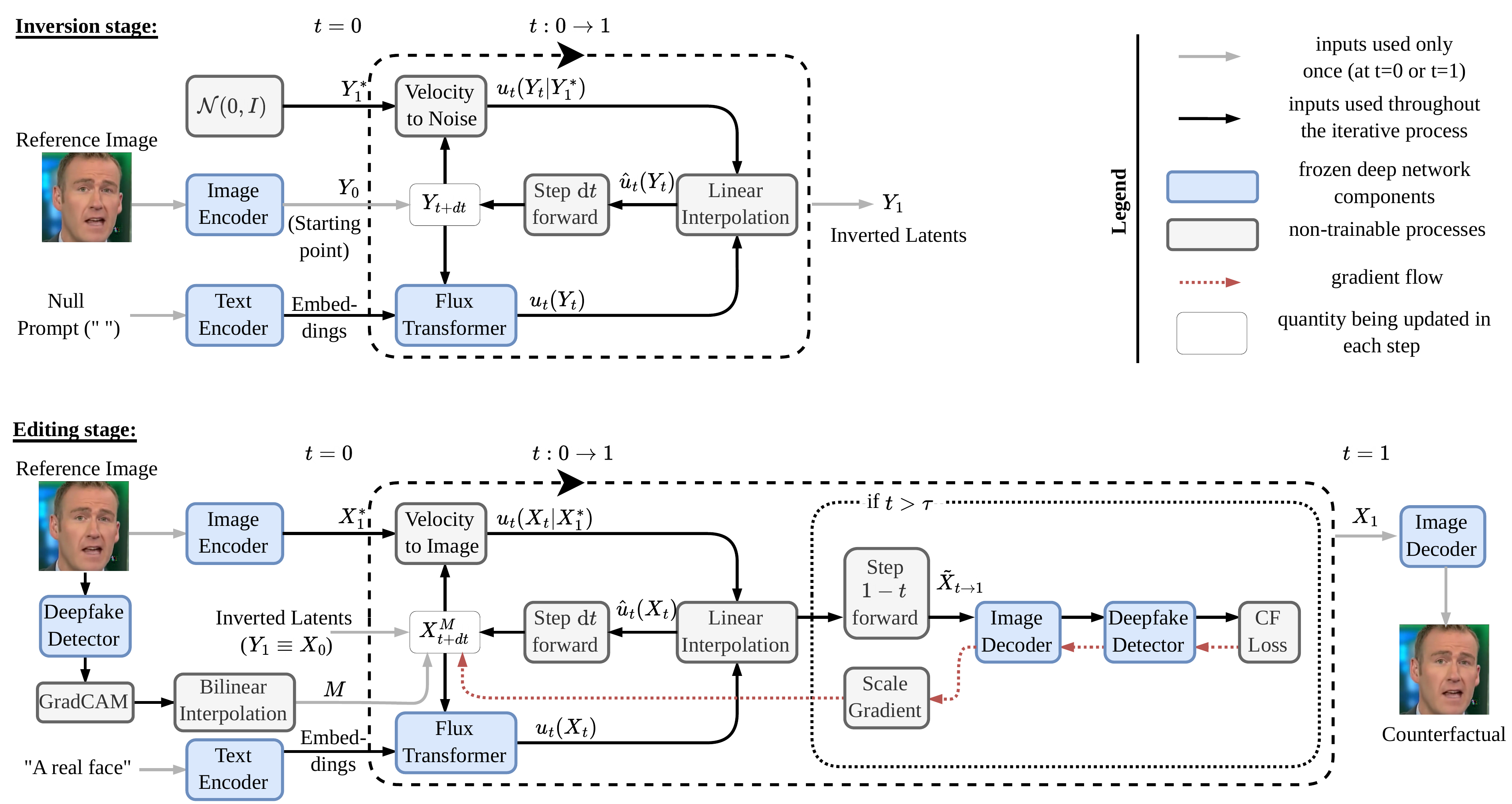}
    \caption{Overview of the proposed Counter-DECEIVR method displaying the inversion and editing stages of the rectified flow method, combined with the classifier guidance phase, and the use of the explanation map.}
    \label{fig:method}
\end{figure*}

Counter-DECEIVR generates a counterfactual example $x'$ for a query image $x$ by editing a rectified flow's latent trajectory so that a target detector $f$ changes its prediction from the query class $c$ to a chosen target class $c'$, while the edit is concentrated on the regions $f$ focuses on to make its prediction. The method has two stages (Fig.~\ref{fig:method}). In the \emph{inversion stage}, we map $x$ to a structured noise latent $y_1$ along a controlled forward ODE (Ordinary Differential Equation), following RF-Inversion \cite{routsemantic2024}. In the \emph{editing stage}, we integrate the reverse ODE from $y_1$ back toward the image manifold, and like in RF-Inversion, this has two phases. In the first phase, the image simply moves back towards the query image. In the second phase, it follows the estimated vector field and, unlike RF-inversion, adds a gradient term at each step to push the trajectory across the detector decision boundary. The central difficulty is that a classifier can only be queried on clean images, whereas the editing trajectory lives in noise. Counter-DECEIVR resolves this by projecting the latent to a clean-image estimate at each guided step, differentiating only through the decoder and detector, and converting the gradient to a vector of a magnitude that is a fraction of the magnitude of the flow step it accompanies. The explanation map of $f$ on $x$ provides a spatial weight that focuses both the field and the gradient on the regions the detector uses.

\subsection{Rectified Flow Inversion}
\label{ssec:prelim}
A rectified flow \cite{lipmanflow2022, liuflow2022} defines a time-varying velocity field $u_t(Y_t)$ whose ODE transports samples between a noise distribution and the data distribution. We parameterize the flow over a continuous time variable $t \in [0,1]$ 

\paragraph{Inversion.} We denote the clean image latent as $Y_0 = \mathcal{E}(x)$, where $\mathcal{E}$ is the image encoder, and sample a noise latent $Y^*_1 \sim \mathcal{N}(0, 1)$. The inversion stage integrates from $t=0$ to $t=1$ the controlled forward ODE, which mixes $u_t(Y_t)$ with a conditional field $u_t(Y_t|Y^*_1)$ pointing toward $Y^*_1$ along the straight-line path: 
\begin{equation}
    \hat{u}_t(Y_t) = (1-\gamma)\cdot u_t(Y_t) + \gamma \cdot u_t(Y_t|Y^*_1)
\end{equation}
With $\gamma=1$, the integration follows the straight-line path towards noise exactly. With $\gamma=0$, it follows the learned field alone.

\paragraph{Editing.} The editing stage runs in the opposite direction. It starts from $Y_1 \equiv X_0$ at $t=0$ and integrates towards an image latent $X_1$ at $t=1$. The learned field is used with the opposite sign ($u_t(Y_t) = -u_{1-t}(X_{1-t})$), and we again mix in a conditional field, here pointing towards the query image latent $X^*_1$
\begin{equation}
    \hat{u}_t(X_t) = (1-\eta)\cdot u_t(X_t) + \eta \cdot u_t(X_t|X^*_1)
    \label{eq:editing}
\end{equation}
With $\eta=1$ the editing trajectory reverses the inversion path exactly and reconstructs the query; with $\eta=0$ it follows the learned reverse field.

\subsection{Guided Rectified Flow Inversion}
\label{ssec:guide}

The inversion stage produces a structured noise latent \(Y_1\) that is independent of the detector \(f\). The editing stage is where the counterfactual intervention occurs. A standard RF-Inversion editing trajectory starts from \(Y_1 \equiv X_0\) and reconstructs the query image. We alter this trajectory to push the image towards the target class \(c'\).

\paragraph{Where to apply guidance.} The detector \(f\) operates on clean images, not noisy latents. Therefore, guidance cannot be applied during the inversion stage, nor can it be applied to the raw output of the velocity field during editing. Instead, we intervene in the second phase of the editing stage.

Let the $N$ editing steps be indexed $k=0,\dots,N-1$, with step $k$ corresponding to time $t=k/N$. Following~\cite{routsemantic2024}, we split the trajectory into two phases at a controller stopping time $\tau\in[0,1]$: the mixing coefficient in Eq.~\eqref{eq:editing} is $\eta_t=\eta$ for $t\le\tau$ and $\eta_t=0$ for $t>\tau$. The first phase reconstructs the query along the exact conditional field toward $X^*_1$; the second follows the learned reverse field, where the trajectory begins to diverge, and the detector guidance is applied. For readability, we report $\tau$ in step units, i.e.\ multiplied by $N$: $\tau=N$ leaves the controller active throughout (pure reconstruction, no guided step), and $\tau=0$ applies guidance from the first step.

\paragraph{Clean-image guidance and gradient bypass.} At each step \(t \ge \tau\), we estimate the final clean image from the current latent \(X_t\) by taking a single forward step to \(t=1\), yielding the projection \(\tilde{X}_{t \to 1}\). We decode this projection into an image, evaluate the detector's cross-entropy loss \(\mathcal{L}_{\text{CF}}\) with respect to the target class \(c'\), and compute the gradient:
\begin{equation}
    g = \nabla_{\tilde{X}_{t \to 1}} \mathcal{L}_{\text{CF}} \left( f \left( \mathcal{D}(\tilde{X}_{t \to 1}) \right), c' \right)
\end{equation}
where \(\mathcal{D}\) is the image decoder. Since the projection is $\tilde{X}_{t \to 1} = X_t + (1 - t)u_t(X_t)$, using $g$ in place of $\nabla_{X_t}\mathcal{L}_{\text{CF}}$ amounts to treating the Jacobian $\partial \tilde{X}_{t \to 1}/\partial X_t$ as the identity, an approximation that becomes exact in the limit $t \to 1$ and, away from that limit, was found effective for rectified flow models by~\cite{patelsteering2024}. Backpropagation therefore flows only through the detector and the decoder, bypassing the rectified flow model and making the per-step cost of guidance independent of the generative model's size.

\paragraph{Gradient scaling.} The magnitude of \(g\) varies significantly across queries. To ensure the intervention is stable, and to minimize the need for hyperparameter tuning, we normalize \(g\) and scale it to a fraction of the magnitude of the current flow step, turning it into a velocity-like vector. The modified latent update is:
\begin{equation}
    X_{t + dt} \leftarrow X_t + \hat{u}_t(X_t) dt - \xi_t \left\| \hat{u}_t(X_t) dt \right\| \frac{g}{\|g\|}
    \label{eq:guide}
\end{equation}
where \(\xi_t\) is a scaling factor decaying over the guided phase.

\paragraph{Masking.} Unconstrained editing creates issues, like affecting irrelevant background details that the second phase of the editing stage cannot reconstruct: doing so requires significantly increasing $\tau$, which in turn makes edits insignificant. We solve the dilemma of needing to significantly edit some areas of the query image, while having to preserve others, by using explanation maps. Specifically, we compute the detector's GradCAM~\cite{selvarajuGradCAMVisualExplanations2017} explanation map for the query image \(x\), apply bilinear interpolation to match the latent resolution, and obtain a spatial mask \(M \in [0,1]^{H \times W}\), broadcast across channels. The mask is applied at two points. First, it blends the vector field: inside the mask, we use the unconstrained flow field \(\hat{u}_t(X_t)\) to allow divergence, while outside the mask, we follow the conditional field \(u_t(X_t|X^*_1)\) to reconstruct the query:
\begin{equation}
    \hat{u}^M_t(X_t) = M \odot \hat{u}_t(X_t) + (1 - M) \odot u_t(X_t|X^*_1)
\end{equation}
Second, it blends the latent update, applying the gradient step within the mask and the reconstruction step outside:
\begin{equation}
\begin{split}
    X_{t + dt} \leftarrow &\; M \odot \left( X_t - \xi_t \left\| \hat{u}_t(X_t) dt \right\| \frac{g}{\|g\|} \right) \\
    &+ (1 - M) \odot X_t +\hat{u}^M_t(X_t) dt
\end{split}
\end{equation}

Because \(M\) is a soft mask, the transition from heavily edited regions to purely reconstructed regions is smooth. The regions that change most are therefore those highlighted by the detector’s explanation map. An explanation map is a poor final explanation because it shows only \emph{where} the detector looks; here it serves as a spatial prior, and the counterfactual provides \emph{what} changes there. The final counterfactual is obtained by decoding the latent at \(t=1\). 
\section{Experiments}
\label{sec:results}
\subsection{Experimental Setup}
\label{ssec:setup}
\begin{table*}
\centering
\caption{Comparison of our method with SoA methods using the FR, COUT, FID, LPIPS, and SSIM metrics. For each method, we select the point on its curve with the highest COUT within $\text{FID}{=}20{\pm}10$ as can be observed from Fig.~\ref{fig:curves}.}
\label{tab:comparison}
\begin{tblr}{
  colspec = {l l l r r r r r}, 
  cell{2}{1} = {r=7}{m},
  cell{2}{2} = {r=4}{m},
  cell{6}{2} = {r=3}{m},
  cell{9}{1} = {r=7}{m},
  cell{9}{2} = {r=4}{m},
  cell{13}{2} = {r=3}{m},
  vline{6} = {1-15}{dashed}, 
}
\toprule
Dataset & Detector & Method & FR↑ & COUT↑ & FID  & LPIPS & SSIM \\
\midrule
FF++ \cite{rosslerfaceforensics2019} & EfficientNet & DISC \cite{thiagarajandesigning2021} & \textbf{1.000} & 0.189 & 14.34 & 0.018 & 0.963 \\
 &  & ACE \cite{jeanneretadversarial2023} & 0.550 & -0.095 & 27.96 & 0.033 & 0.960 \\
 &  & Stable Diffusion V1.4 \cite{yangunveiling2025} & 0.983 & 0.437 & 18.73 & 0.053 & 0.913 \\
 &  & Counter-DECEIVR & 0.998 & \textbf{0.815} & 18.65 & 0.036 & 0.963 \\
\midrule
 & Swin V2 & ACE \cite{jeanneretadversarial2023} & 0.500 & -0.089 & 18.81 & 0.020 & 0.971 \\
 &  & Stable Diffusion V1.4 \cite{yangunveiling2025} & \textbf{0.955} & 0.729 & 19.37 & 0.043 & 0.939 \\
 &  & Counter-DECEIVR & 0.950 & \textbf{0.754} & 20.75 & 0.029 & 0.975 \\
\midrule
Celeb-DF \cite{licelebdf2020} & EfficientNet & DISC \cite{thiagarajandesigning2021} & \textbf{1.000} & 0.457 & 12.49 & 0.012 & 0.975 \\
 &  & ACE \cite{jeanneretadversarial2023} & 0.500 & -0.105 & 21.04 & 0.016 & 0.980 \\
 &  & Stable Diffusion V1.4 \cite{yangunveiling2025} & 0.948 & 0.298 & 27.45 & 0.068 & 0.899 \\
 &  & Counter-DECEIVR & \textbf{1.000} & \textbf{0.607} & 20.37 & 0.029 & 0.964 \\
\midrule
 & Swin V2 & Stable Diffusion V1.4 \cite{yangunveiling2025} & 0.982 & 0.802 & 27.43 & 0.045 & 0.945 \\
 &  & ACE \cite{jeanneretadversarial2023} & 0.650 & 0.185 & 12.64 & 0.012 & 0.984 \\
 &  & Counter-DECEIVR & \textbf{1.000} & \textbf{0.990} & 22.94 & 0.028 & 0.954 \\
\bottomrule

\end{tblr}
\end{table*}

\begin{figure*}[htbp]
    \centering  
    \includegraphics[width=1\linewidth]{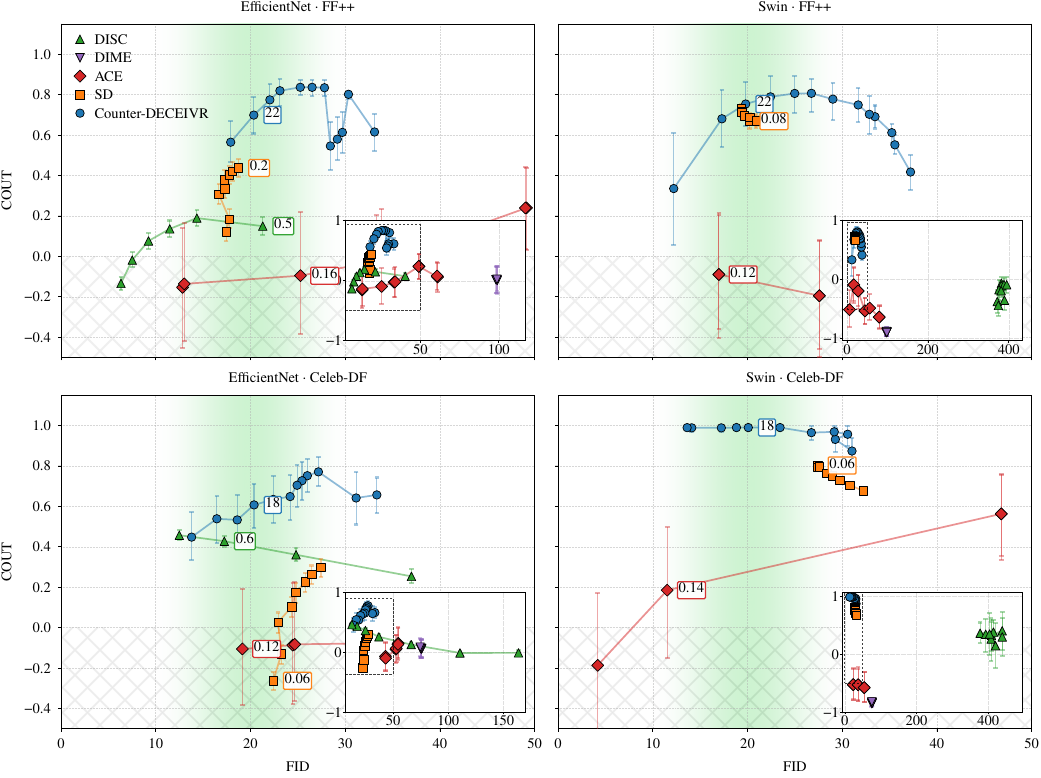}
    \caption{COUT as a function of FID for each method, obtained by sweeping editing strength. Higher is better. Each marker is one operating point; the shaded region marks $\text{FID}=20\pm10$, from which the operating point reported in Table~\ref{tab:comparison} is selected. The inset shows the full range of FID values attained by each method, including methods whose entire curve lies above $\text{FID}{=}50$.}
    \label{fig:curves}
\end{figure*}

\begin{figure*}[htbp]
    \centering  
    \includegraphics[width=0.84\linewidth]{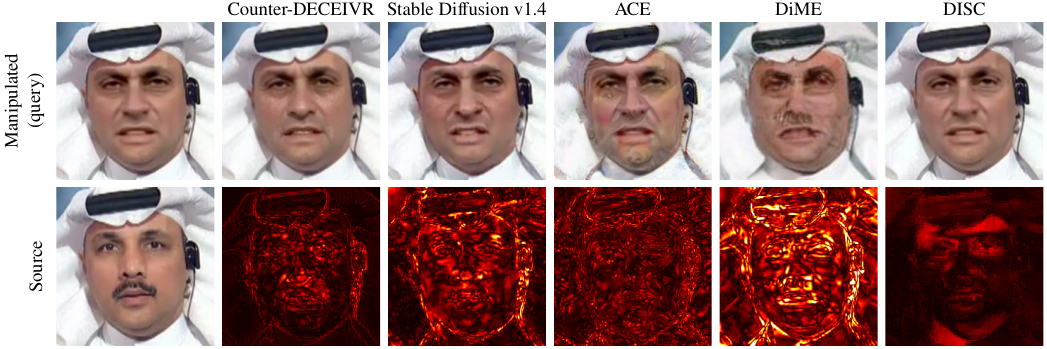}
    \caption{Qualitative comparison of Counter-DECEIVR and the baselines on the same queries. Each row shows the query and the counterfactual produced by each method, with the difference between query and counterfactual as an inset.}
    \label{fig:methods}
\end{figure*}

\begin{table*}
\centering
\caption{Ablation study of explanation map masking (Grad-CAM) with no masking.}
\label{tab:ablation}
\begin{tblr}{
  vline{4} = {1-4}{dashed},
}
\toprule
Masking & FR↑ & COUT↑ & FID  & LPIPS & SSIM \\
\midrule
No Masking & \textbf{1.000}	& 0.617 & 21.25 & 0.057 & 0.866 \\
Proposed Grad-CAM-based Masking  & 0.998 & \textbf{0.815}  & 18.65 & 0.036 & 0.963 \\
\bottomrule

\end{tblr}
\end{table*}

\begin{figure}[htbp]
    \centering  
    \includegraphics[width=1\linewidth]{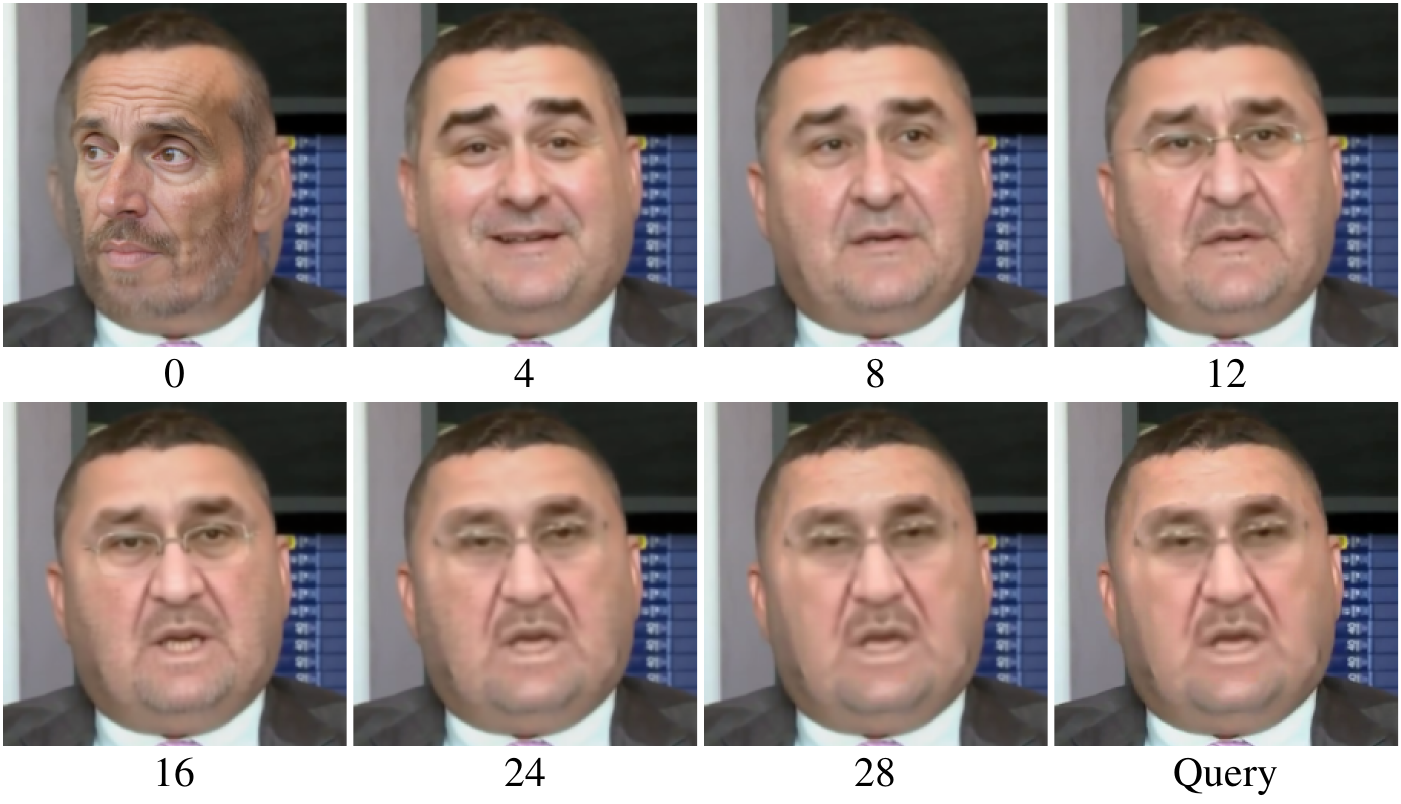}
    \caption{Effect of $\tau$ on Counter-DECEIVR's counterfactuals for a fixed query. As $\tau$ decreases, the edit becomes more drastic, while the background remains intact. For small $\tau$ the masked region is largely resampled from the flow prior, so changes such as the removed glasses reflect the prior rather than detector evidence. The sweep traces the trade-off quantified in
 Figure~\ref{fig:curves}.}
    \label{fig:tau}
\end{figure}

We evaluate on two deepfake detectors: an EfficientNet-based detector trained on the FF++~\cite{rosslerfaceforensics2019} dataset that classifies an image as Real or as one of four manipulation methods, and a Swin-v2-based binary detector also trained on FF++. Each counterfactual generation method is evaluated with both detectors on FaceForensics++~\cite{rosslerfaceforensics2019} and Celeb-DF~\cite{licelebdf2020}. Counterfactuals are generated for, and evaluated against, a single detector; since detectors rely on different evidence, transfer is neither expected nor desired. All methods are evaluated on the same fixed query set: for each dataset we sample 400 manipulated test images, disjoint from any data used to train or fine-tune a baseline. For the EfficientNet detector, the query class is the manipulation type the detector predicts, and the target class is Real; for the Swin-v2 detector, the query class is Fake, and the target class is Real. We consider fake$\rightarrow$real only: it exposes the evidence behind a detection, whereas real$\rightarrow$fake is satisfied by introducing any artifact. A query is included only if the detector under evaluation classifies it correctly, so that every counterfactual is asked to overturn an actual detection rather than an existing error. For the rectified flow model, we use FLUX.1-dev~\cite{labs2025flux1kontextflowmatching}, following~\cite{routsemantic2024}. Both the inversion and editing stages use 28 steps, again following~\cite{routsemantic2024}, with $\gamma{=}0.5$ during inversion; during editing $\eta{=}1.0$ before step $\tau$ and $\eta{=}0$ from $\tau$ onward. Inversion uses a null prompt; editing uses the prompt ``A real face'', the same prompt used by~\cite{yangunveiling2025}, with guidance scale $3.5$, the default value for FLUX.1-dev. The guidance scale $\xi_t$ of Eq.~(\ref{eq:guide}) starts at $0.1$ and decays geometrically to $0.05$ over the guided steps. Unlike a pixel-space attack, the gradient is never applied to the image: it perturbs the latent trajectory, and the remaining flow steps and the decoder map the result back to the image manifold, while $\xi_t$ decays so that the final steps are dominated by the flow. GradCAM~\cite{selvarajuGradCAMVisualExplanations2017} is computed for the query class on the last feature map before the classification head; for Swin-v2, whose final stage outputs tokens rather than a feature map, we reshape them to their $7\times7$ spatial grid. Images are generated at $1024\times1024$ and resized to $224\times224$ or $256\times256$, the input resolution each detector was trained at; this resize sits inside the gradient path, so gradients reach the latent through it. The map is then converted to a soft per-pixel weight in $[0,1]$ by histogram equalization and a $\gamma$ correction that fixes the mean intensity at $0.5$, so that guidance strength reflects the relative ordering of the CAM rather than its raw scale. The explanation map is resized to the latent resolution with bilinear interpolation. $\tau$ is the only parameter we vary across experiments.

We evaluate counterfactuals along the three requirements of Section~\ref{sec:related} (minimality, saliency, and plausibility) using five metrics. COUnterfactual Transition (COUT$\uparrow$) scores \cite{khorramcycleconsistent2022} measure how efficiently a counterfactual increases the detector's confidence in the target class, while decreasing its confidence in the query class. Pixels are ranked by their absolute difference $|x'-x|$ between the counterfactual and query and inserted from $x'$ into $x$ in $T{=}10$ ordered steps. With $f_k(\check{x}_{(t)})$ the detector's score for class $k$ at insertion step $t$, we compute $\mathrm{AUPC}_k=\frac{1}{T}\left[\frac{1}{2}f_k(\check{x}_{(0)})+\sum^{T-1}_{t=1}f_k(\check{x}_{(t)}) + \frac{1}{2}f_k(\check{x}_{(T)})\right]$ for the query class $c$ and the target class $c'$, and report $\text{COUT} = \mathrm{AUPC}_{c'} - \mathrm{AUPC}_c\in[-1,1]$. COUT rewards edits that flip the decision early in the insertion order, i.e., sparse changes confined to the pixels the detector weighs most; edits that require replacing most of the image score poorly even when they are valid. Because the ordering uses $|x' -x|$ for every method, no method is advantaged by its own localization: COUT captures the sparsity of the edit and, in part, its alignment with detector-sensitive pixels. Alongside COUT, we report the Flip Rate (FR), the percentage of cases in which the counterfactual actually flips the detector's original prediction.

We complement COUT and FR with three fidelity metrics that measure how much the counterfactual diverges from its query. FID is the Fr\'echet distance between the feature distributions of the query set and the counterfactual set. SSIM and LPIPS measure structural and perceptual similarity between each query and its counterfactual. These three metrics are not monotone in the counterfactual setting: each has an optimum rather than a preferred direction, since values indicating no difference mean no edit was made, while values indicating a large difference mean the image was destroyed. No single operating point is therefore directly comparable across methods, because any method can trade COUT against fidelity by editing more or less aggressively. We accordingly compare methods as curves rather than at a single point. SSIM and LPIPS are poorly suited to this comparison: both are pairwise, pixel-level similarities that saturate, so a small but decisive edit barely moves them, while a large but semantically irrelevant one moves them substantially. FID, as a distributional distance between the query and counterfactual sets, responds smoothly across that range and provides a finer-grained axis. We therefore sweep the editing strength of each method ($\tau$ for Counter-DECEIVR, $\kappa$ for DISC \cite{thiagarajandesigning2021}, the diffusion strength for ACE \cite{jeanneretadversarial2023}, DiME \cite{jeanneretdiffusion2022} and the trained Stable Diffusion v1.4 \cite{yangunveiling2025}) and report COUT as a function of FID, tracing the trade-off curve each method can achieve. A method that attains higher COUT at equal FID is preferred.

\subsection{Quantitative Comparisons}
\label{ssec:quant}
We compare against four counterfactual generation methods: DISC \cite{thiagarajandesigning2021}, DiME \cite{jeanneretdiffusion2022}, ACE \cite{jeanneretadversarial2023}, and a Stable Diffusion v1.4 model finetuned on deepfake datasets by \cite{yangunveiling2025}. No implementation was released for DISC \cite{thiagarajandesigning2021}, so we follow the authors' description: we train the auxiliary error-prediction module \cite{yoolearning2019} on the FF++ dataset \cite{rosslerfaceforensics2019} in a post-hoc manner using the same detectors, and, following the findings of \cite{thiagarajandesigning2021}, use INR \cite{sitzmannimplicit2020} as the representation network rather than DIP \cite{ulyanovdeep2020}. DiME and ACE both require a diffusion model trained on the target data, which we train on the FF++ dataset split. No implementation was released for \cite{yangunveiling2025} either, so we follow the authors' description and finetune Stable Diffusion v1.4 on the FF++ dataset jointly with each detector. Unlike the original authors, we use a single detector rather than an ensemble, so that each counterfactual is specific to one detector. We provide our implementations of DISC and the Stable Diffusion v1.4 guided diffusion training together with our method. These adaptations train every baseline on the target distribution: DISC trains an error-prediction module, DiME and ACE train a diffusion model, and the Stable Diffusion baseline finetunes the generator with the detector gradient. Counter-DECEIVR is not trained on the target distribution. Both it and the Stable Diffusion baseline start from web-scale pretraining, but only the baseline is then trained further: it is fine-tuned on the target distribution with the detector gradient, whereas Counter-DECEIVR uses an off-the-shelf detector and the pretrained rectified flow model as they are.

Figure~\ref{fig:curves} traces, for each method, the COUT it attains as a function of the FID of its counterfactuals. The curves characterise the trade-off each method can reach rather than a single operating point, since any method can raise COUT by editing more aggressively, at the cost of FID. Counter-DECEIVR's curve lies to the upper left of all baselines across the plotted range: at equal FID it attains higher COUT, and at equal COUT it attains lower FID. DISC approaches Counter-DECEIVR at lower FID in the EfficientNet - Celeb-DF case.

Table~\ref{tab:comparison} reports FR and COUT, as well as SSIM and LPIPS, at a matched operating point. For each method, we select the point on its curve with the highest COUT within $\text{FID}{=}20{\pm}10$, the range we found to correspond to counterfactuals that make a visible edit while preserving the query's identity. A method with no point below $\text{FID}=50$ is excluded from the table, since it cannot produce a counterfactual that stays near the query distribution at any editing strength; its full curve is still shown in Figure~\ref{fig:curves}. DISC is reported for EfficientNet only. Its objective balances the classifier loss against feature-space distance terms; due to the nature of the Swin-v2 architecture, features are much smaller, and the optimization does not reach a stable reconstruction of the query image without layer-specific feature rescaling, which would depart further from the published method. DiME is excluded from the table, since no operating point below the $\mathrm{FID}=50$ cutoff point was observed.

\subsection{Qualitative Results}
\label{ssec:qual}

Figure~\ref{fig:methods} compares counterfactuals from each method on the same queries, alongside the difference between the query and the counterfactual images. The qualitative difference is in where the edit lands. Counter-DECEIVR's edits concentrate on the regions the explanation map highlights, leaving the background and the rest of the face largely intact, so the change can be read as evidence for the detector's decision. 

\subsection{Ablations}
\label{ssec:abl}
Table~\ref{tab:ablation} compares the masking strategy, holding all other parameters fixed: (i) no mask, where guidance is applied uniformly across the image; (iii) soft weighting by the Grad-CAM \cite{selvarajuGradCAMVisualExplanations2017} explanation map, as in our default configuration. Without masking, COUT falls to $0.617$ while FID rises to $21.249$ and SSIM falls to $0.866$: the detector is flipped in $100\%$ of cases by editing broadly, and the resulting change no longer identifies detector evidence.

Figure~\ref{fig:tau} sweeps $\tau$ from 0 to 28. At $\tau{=}28$ no guided step is taken, and Counter-DECEIVR reduces to RF-Inversion reconstruction; the flip rate there is $0\%$, so flips at lower $\tau$ come from the guided edit, not from re-encoding. As $\tau$ decreases, more and more edits are made to the image; remarkably, even at $\tau=0$, fine details of the background remain intact thanks to our masking strategy. Together with the quantitative curves in Figure~\ref{fig:curves}, this illustrates that a single parameter moves Counter-DECEIVR along its trade-off frontier, so that the operating point used in Table~\ref{tab:comparison} can be selected per application, or per case, at inference time.

\subsection{Limitations}
\label{ssec:limit}
Counter-DECEIVR requires gradient access to the detector and cannot be run against detectors exposed only through an API. For open detectors, a white-box adversarial attack is cheaper and evades detection; Counter-DECEIVR's advantage is that the resulting change is tied to the detector's evidence, but it offers no advantage over existing white-box attacks, limiting its misuse potential.
\section{Conclusion}
\label{sec:conclusion}
We introduced Counter-DECEIVR, a training-free framework for generating counterfactual explanations for deepfake detectors. By combining rectified flow inversion with detector-guided editing, Counter-DECEIVR avoids training noise-aware classifiers and concentrates edits to salient regions using the detector's own explanation map. Experiments across two detectors and datasets show that Counter-DECEIVR achieves a superior trade-off between counterfactual transition and fidelity compared to training-based baselines. Future directions of this work could extend its applicability to video deepfakes.

{
    \small
    \bibliographystyle{ieeenat_fullname}
    \bibliography{main.bib}
}

\end{document}